\pdfoutput=1

\documentclass[11pt]{article}

\usepackage{EMNLP2022}

\usepackage{times}
\usepackage{latexsym}

\usepackage[T1]{fontenc}

\usepackage[utf8]{inputenc}

\usepackage{microtype}

\usepackage{inconsolata}

\usepackage{hyperref}       
\usepackage{url}            
\usepackage{booktabs}       
\usepackage{nicefrac}       
\usepackage{bm}

\usepackage{times}
\usepackage{latexsym}

\usepackage{algorithm}
\usepackage{algpseudocode}
\usepackage{amsmath}
\usepackage{bbm}
\usepackage{CJK}
\usepackage{graphicx}
\usepackage{float}
\usepackage{xcolor,colortbl}
\usepackage{bigstrut,bigdelim}
\usepackage{paralist}
\usepackage{diagbox}
\usepackage{wrapfig}
\usepackage{verbatim}
\usepackage{subfigure}
\usepackage{multicol}
\usepackage{multirow}
\usepackage{makecell}
\usepackage{amssymb}
\usepackage{xcolor}

\DeclareMathOperator*{\argmax}{arg\,max}

\title{Collaborative Reasoning on Multi-Modal Semantic Graphs for Video-Grounded Dialogue Generation}

\author{
    Xueliang Zhao$^{1,2}$\footnotemark[1], Yuxuan Wang$^{1,2}$\footnotemark[1], Chongyang Tao$^1$, \\
    \textbf{Chenshuo Wang}$^{1,2}$ \and
    \textbf{Dongyan Zhao}$^{1,2,3}$\footnotemark[2] \\
    $^1$Wangxuan Institute of Computer Technology, Peking University\\
    $^2$Center for Data Science, AAIS, Peking University\\
    $^3$Beijing Institute for General Artificial Intelligence \\
    \texttt{\{xl.zhao,chongyangtao,zhaody\}@pku.edu.cn} \quad
    \texttt{\{wyx,wcs\}}@stu.pku.edu.cn 
}

\begin{document}
\maketitle

\renewcommand{\thefootnote}{\fnsymbol{footnote}}
\footnotetext[1]{Equal Contribution.}
\footnotetext[2]{Corresponding author: Dongyan Zhao.}
\setcounter{footnote}{0}
\renewcommand{\thefootnote}{\arabic{footnote}}

\begin{abstract}

We study video-grounded dialogue generation, where a response is generated based on the dialogue context and the associated video. The primary challenges of this task lie in (1)  the difficulty of integrating video data into pre-trained language models~(PLMs) which presents obstacles to exploiting the power of large-scale pre-training; and (2) the necessity of taking into account the complementarity of various modalities throughout the reasoning process. Although having made remarkable progress in video-grounded dialogue generation, existing methods still fall short when it comes to integrating with PLMs in a way that allows information from different modalities to complement each other. To alleviate these issues, we first propose extracting pertinent information from videos and turning it into reasoning paths that are acceptable to PLMs. Additionally, we propose a multi-agent reinforcement learning method to collaboratively perform reasoning on different modalities~(i.e., video and dialogue context). Empirical experiment results on two public datasets indicate that the proposed model can significantly outperform state-of-the-art models by large margins on both automatic and human evaluations.

\end{abstract}
\section{Introduction}

Conversing with computers has become a crucial step toward general artificial intelligence, and it has attracted increasing attention from AI and NLP researchers.
Multi-turn dialogue response generation and multi-modal question answering are two high-profile initiatives made toward this goal.
The task of multi-turn dialogue response generation necessitates the agent comprehending the key information in the dialogue context in order to provide a cohesive, fluent and informative response~\cite{zhao2017learning,tao2018get}.
Multi-modal question answering, on the other hand, necessitates the agent's understanding of both the textual and visual contexts~\cite{antol2015vqa,tapaswi2016movieqa,jang2017tgif}.
The video-grounded dialogue~\cite{alamri2018audio,pasunuru2018game} is a generalization of the above two tasks, in which the agent must observe multi-modal contents and engage in a conversation with the human, rather than simply responding to the last utterance or ignoring the visual contents.
Compared to multi-turn dialogue response generation and multi-modal question answering, the distinctive challenges posed by video-grounded dialogue generation can be summarized as:
(1) Unlike traditional multi-turn dialogue that can directly use large-scale pre-trained language models~(PLMs), video-grounded dialogue cannot directly use PLMs due to their incapacity to process video input;
(2) In comparison to multi-modal question answering, video-grounded dialogue necessitates reasoning on both video and multi-turn textual context, and there is usually a complementarity between different modalities that should be taken into account.

Although having made notable progress in video-grounded dialogue, existing approaches still fail to recognize the aforementioned challenges.
On one hand, existing approaches cannot be effectively combined with PLMs, which presents obstacles to exploiting the power of state-of-the-art pre-training technology. 
The reasons can be summarized into two categories:
(1) Simply appending the video features to the text embeddings presents a challenge for the model to obtain an in-depth understanding of the video~\cite{li2020bridging,le2020video,le2021learning}.
To investigate this problem further, we compare the performance of these models before and after removing the video from the input. As demonstrated in Table~\ref{tab:exp-pilot}, most metrics only show a tiny shift, and several even increase once the video is removed;
and (2) Overly complex designs for the Transformer that are difficult to transfer to PLMs~\cite{le2020bist,Kim2021StructuredCG,Geng2021DynamicGR}.
On the other hand, multi-modal information should be used in conjunction with each other, and reasoning on different modalities should be done \textbf{collaboratively} rather than \textbf{independently}.
Existing approaches fall short when it comes to reasoning jointly on multi-modalities, since they either separate the reasoning of different modalities~\cite{li2020bridging} or employ a cross-modal attention mechanism which is difficult to train without direct supervision~\cite{le2020bist,Kim2021StructuredCG,Geng2021DynamicGR}.

\begin{table}[t!]
\centering
\resizebox{0.95\linewidth}{!}{
\begin{tabular}{lcccc}
\toprule
\multicolumn{1}{c}{Model}    & \multicolumn{1}{c}{BLEU4} & \multicolumn{1}{c}{METEOR} & \multicolumn{1}{c}{ROUGE-L} & CIDEr \\ \midrule
\multicolumn{5}{c}{\textit{with} video}                                                                       \\  \midrule
\multicolumn{1}{l}{RLM}      & \multicolumn{1}{c}{0.402} & \multicolumn{1}{c}{0.254}  & \multicolumn{1}{c}{0.544}   & 1.052 \\ 
\multicolumn{1}{l}{VGD-GPT2} & \multicolumn{1}{c}{0.388} & \multicolumn{1}{c}{0.251}  & \multicolumn{1}{c}{0.539}   & 0.998 \\ 
\multicolumn{1}{l}{PDC-GPT}  & \multicolumn{1}{c}{0.385} & \multicolumn{1}{c}{0.260}  & \multicolumn{1}{c}{0.545}   & 1.010 \\ 
\multicolumn{1}{l}{Ours}     & \multicolumn{1}{c}{0.414} & \multicolumn{1}{c}{0.265}  & \multicolumn{1}{c}{0.558}   & 1.078 \\ \midrule
\multicolumn{5}{c}{\textit{w/o} video}                                                                        \\ \midrule
\multicolumn{1}{l}{RLM}      & \multicolumn{1}{c}{0.401} & \multicolumn{1}{c}{0.255}  & \multicolumn{1}{c}{0.545}   & 1.038 \\
\multicolumn{1}{l}{VGD-GPT2} & \multicolumn{1}{c}{0.393} & \multicolumn{1}{c}{0.251}  & \multicolumn{1}{c}{0.537}   & 1.016 \\
\multicolumn{1}{l}{PDC-GPT}  & \multicolumn{1}{c}{0.388} & \multicolumn{1}{c}{0.261}  & \multicolumn{1}{c}{0.543}   & 1.020 \\
\multicolumn{1}{l}{Ours}     & \multicolumn{1}{c}{0.405} & \multicolumn{1}{c}{0.264}  & \multicolumn{1}{c}{0.554}   & 1.064 \\ 
\bottomrule
\end{tabular}
}
\caption{Pilot study on AVSD$@$DSTC7. }
\vspace{-2mm}
\label{tab:exp-pilot}
\end{table}

To address the aforementioned issues, we propose extracting relevant information from videos and converting it into reasoning paths, which are in the form of natural language and can be fed directly into PLMs. Besides, we propose a multi-agent reasoning framework that is based on the multi-agent reinforcement learning~(MARL) theory. Specifically, we design a video agent and a context agent which learn to find the chains of reasoning on the multi-modal semantic graphs. We further design a central communicator to make the two agents work in a collaborative manner. Our framework has the following advantages: (1) the multi-modal reasoning paths are compatible with the input of PLMs; (2) the reasoning process can be ``supervised'' by designing appropriate reward functions; and (3) the communication mechanism allows the information from different modalities better complement each other. We conduct extensive experiments on two benchmark datasets for video-grounded dialogue generation, including AVSD$@$DSTC7~\cite{alamri2018audio} and Twitch-FIFA~\cite{pasunuru2018game}. Experiment results show that, thanks to the multi-agent reasoning framework, our model can significantly outperform state-of-the-art methods in terms of both automatic and human evaluations.

Our contributions in the paper are three-fold:
(1) Identifying the issue that current PLMs-based approaches are unable to fully comprehend the video content although showing promising results in automatic evaluation metrics.
(2) Proposal of a multi-agent reasoning framework upon PLMs that can let information from different modalities reinforce each other and discover multi-modal reasoning paths.
(3) Empirical verification of the effectiveness of the proposed model on two benchmarks of video-grounded dialogue generation.

\section{Related Work}
The majority of early works on dialogue generation use hand-crafted rules or templates to construct dialogue systems~\cite{weizenbaum1966eliza,Wallace2009TheAO}. A number of initiatives have been made to develop end-to-end open-domain dialogue generation models~\cite{ritter-etal-2011-data,10.5555/3305381.3305510,NIPS2017_3f5ee243}, which have been inspired by the developments in the field of machine translation. Following that, the vanilla encoder-decoder architecture is frequently utilized to enhance response quality, and numerous modifications to this architecture have been made to enhance response diversity~\cite{zhao2017learning,tao2018get}, model the structure of conversation contexts~\cite{zhang-etal-2019-recosa}, introduce external knowledge~\cite{dinan2018wizard,zhao2020knowledge} and control response attributes~\cite{Wang2018LearningTA,See2019WhatMA,Wang2020CuewordDN}.

The research on generating dialogue from video was started by \citet{alamri2018audio}. After that, \citet{hori2019end} present an LSTM-based encoder-decoder architecture with multi-modal attention that merely combines textual and visual data via a projection matrix. A multi-modal transformer network is introduced in \citet{Le2019MultimodalTN} to encode videos and incorporate data from several modalities. \citet{hori2019joint} uses a joint student-teacher learning approach to make up for a missing video description in which the student network is trained to mimic the teacher's response. VGD-GPT~\cite{le2020video} is based on a pre-trained GPT-2 model and formulates the video-grounded dialogue generation as a sequence-to-sequence task. On a pre-trained GPT-2 model, RLM~\cite{li2020bridging} provides a multi-task learning strategy. Additionally, BiST~\cite{le2020bist} models the dependencies between text and visual in two directions: spatial to temporal and temporal to spatial. With visual attention, PDC-GPT~\cite{le2021learning} learns to anticipate the reasoning process on turn-level semantic graphs. For further reasoning, SCGA~\cite{Kim2021StructuredCG} constructs a structured graph based on a multi-modal coreference technique, while STSGR~\cite{Geng2021DynamicGR} introduce a shuffled transformer reasoning framework on semantic scene graph. In contrast to previous approaches, this paper focuses on how to build a multi-modal reasoning approach that can cooperate with PLMs in a way that facilitates the complementary nature of information from various modalities.

The study of reasoning on various types of graph structures for dialogue generation is related to our work. \citet{moon2019opendialkg} create a KG walk path for each entity retrieved in an effort to explain conversational reasoning in a natural way. \citet{Jung2020AttnIOKG} develop a dialogue-conditioned path traversal model with attention flows and improve the comprehension of the path reasoning process. \citet{Xu2020ConversationalGG} propose to represent dialogue transitions as graphs. Previous approaches typically concentrate on textual graphs, but video-grounded dialogue contains multi-modal contexts, which makes it difficult to conduct reasoning.

\begin{figure*}
\centering
\includegraphics[width=0.9\textwidth]{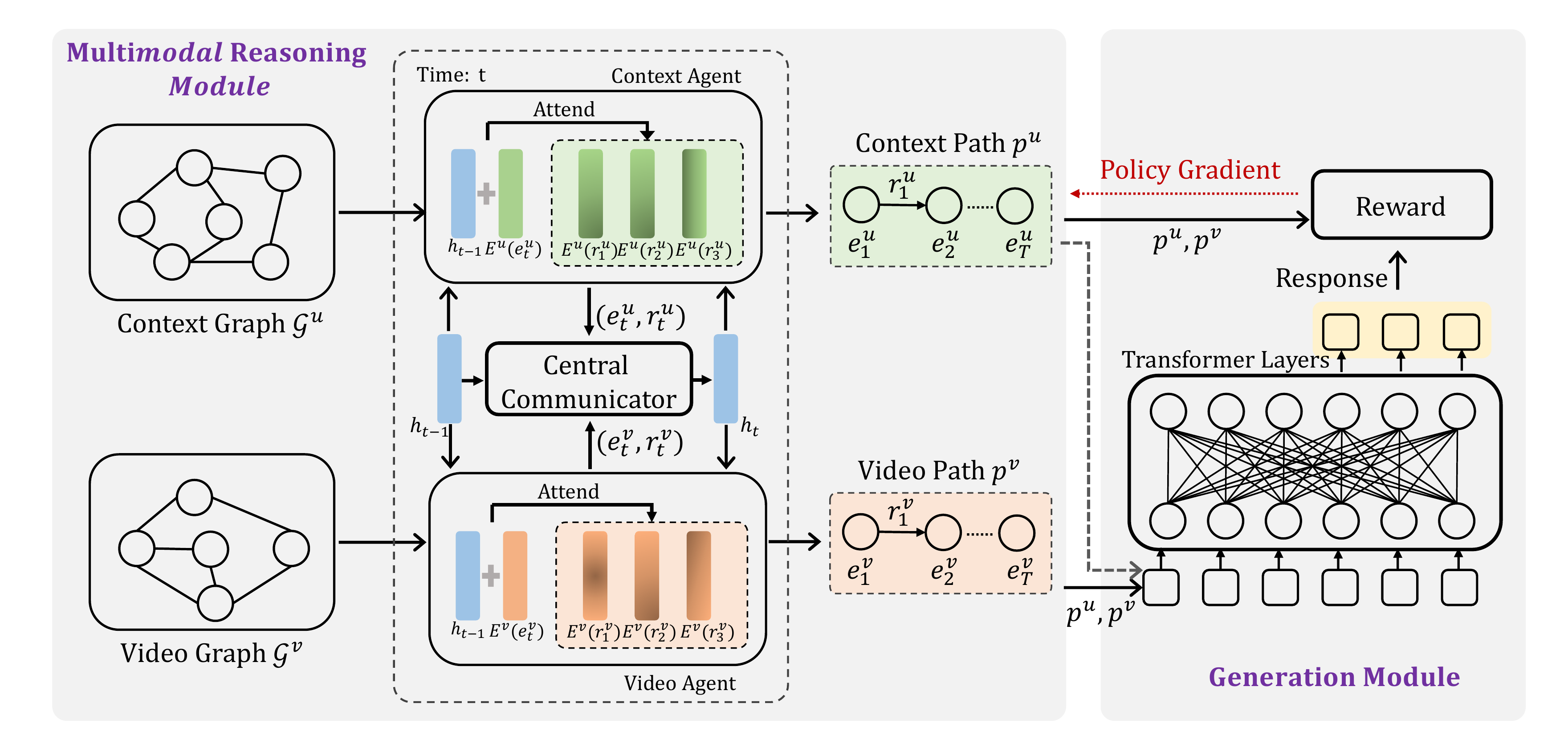}
\caption{Architecture of the proposed model.}
\label{fig:model}
\vspace{-2mm}
\end{figure*}

\section{Approach}

\subsection{Overview}
\label{sec:overview}

Suppose that we have a dataset $\mathcal{D} = \{V_i, U_i, R_i\}_{i=1}^{N}$ with $N$ denoting the total number of datapoints. For the $i$-th datapoint, $V_i$ signifies a brief video clip, $U_i=\{u_{i,1},u_{i,2},\cdots,u_{i,n}\}$ serves as the dialogue context with $u_{i,j}=\{w_{i,j}^{1}, w_{i,j}^{2}, \cdots, w_{i,j}^{m}\}$ denoting the $j$-th utterance. 
$n$ and $m$ are the number of utterances in a context and the number of words in an utterance respectively.
$R_i$ is a response that is factually consistent with the video while also catching up with the dialogue context.
Our goal is to learn a generation model $p(R|V,U;\theta)$\footnote{We omit the subscript to reduce clutter.} ($\theta$ denotes the parameters of the model) from $\mathcal{D}$, so that given a new dialogue context $U$ associated with a video $V$, one can generate a response following $p(R|V,U;\theta)$.

To alleviate the heterogeneity of different modalities, we first represent the video as well as the dialogue context as semantic graphs~(will be elaborated in Section~\ref{sec:graph}).
Figure~\ref{fig:model} illustrates the architecture of the proposed model. In a nutshell, the model is composed of a multi-modal reasoning module and a generation module. The multi-modal reasoning module is responsible for extracting crucial signals from multi-modal contexts~(Section~\ref{sec:reasoning}). Specifically, it consists of a video agent, a text agent and a central communicator. 
The video agent and the text agent are responsible for extracting reasoning paths from the video semantic graph and the text semantic graph respectively. Taking the latest context utterance as input, they determine the query entities from which they start traversing the graphs to find the answer-providing paths.
To search for answer-providing paths more efficiently, we devise a central communicator to transport the entire path histories between video and text agents.
The reasoning paths, which form interpretable provenances for the prediction, are integrated by the generation module to synthesize a response~(Section~\ref{sec:generation}).

\subsection{Multi-Modal Graph Construction}
\label{sec:graph}

The crucial step in building the semantic graph for video reasoning is gathering the collection of facts from the unstructured video data, which take the form of subject-predicate-object triplets.
Although there have been some previous attempts to extract such triplets from videos using relation detection~\cite{liu2020beyond}, the models that have been made public struggle to build the proper relations because of the dramatic domain discrepancy between their training corpus and the benchmark dataset for video-grounded dialogue.
Therefore, we resort to video action recognition~\cite{zhu2020comprehensive} to extract meaningful structural representations from video.
Specifically, we first employ the slowfast model~\cite{feichtenhofer2019slowfast}, which is pre-trained on the Charades ~\cite{sigurdsson2016hollywood} and Kinetics dataset~\cite{kay2017kinetics}, to extract all potential action classes and only reserve those with a probability greater than $0.5$.
Given the extracted facts $\{(e^v_s, r, e^v_o)\}$ with $e^v_s$, $r^v$ and $e^v_o$ standing for subject, predicate and object respectively, we can construct a video semantic graph $\mathcal{G}^v=(N^v, E^v)$ in which the entities $e^v_s$ and $e^v_o$ are represented as nodes~(i.e., $e^v_s, e^v_o \in N^v$) and the relation $r^v$ is represented as a labeled edge connecting them~(i.e., $(e^v_s, r, e^v_o) \in E^v$).

The semantic graph for dialogue context, $\mathcal{G}^u=(N^u, E^u)$, is constructed in a similar way, except that we employ open information extraction~(OpenIE) technology to extract subject-predicate-object triplets.
Specifically, we first apply the co-reference resolution tool~(e.g., AllenNLP~\cite{Gardner2017AllenNLP} in our experiments) to restore all the pronouns to their original name entities. 
Then we extract all relation triplets in a dialogue context by combining the outputs of OpenIE 5.1~\cite{Saha2018OpenIE} and Stanford OpenIE~\cite{angeli2015leveraging}. 
We further remove unnecessary information after getting all the triplets by combining all entities with high semantic similarity, as determined by the cosine similarity
between the word2vec embeddings~\cite{mikolov2013distributed}.

\subsection{Multi-Agent Reasoning Process}
\label{sec:reasoning}

Inspired by recent advances in graph-grounded generation~\cite{moon2019opendialkg,Xu2020ConversationalGG}, we decompose the problem of video-grounded generation into two steps: (1) identify answer-providing paths on the graph that might contain crucial signals for catching up with the context; (2) generate a response using the extracted paths as additional information.
However, independently extracting the chains of reasoning for each modality will result in a sub-optimal solution, since the video provides crucial guidance for text reasoning and vice versa.
To this purpose, we propose a multi-agent reasoning framework, where agents responsible for different modalities can work in a collaborative manner.

We formulate the multi-modal reasoning task as a partially observable multi-agent sequential decision process on semantic graphs $\mathcal{G}^v$ and $\mathcal{G}^u$.
Intuitively, we want a state $s_t$ at time $t$ to be a summary of previous experiences: $s_t=(o_1, a_1, \cdots, a_{t-1}, o_t)$, where $o_t=(e_t^v, e_t^u)$ and $a_t=(r_t^v, r_t^u)$ stand for observations/entities of all agents at time $t$ and actions/relations taken by them respectively. $o_1=(e_1^v, e_1^u)$ is the query entity to start traversing on the graphs and is defined as:
\begin{equation}
\begin{aligned}
    e_1^v &= \argmax_{e^v \in N^v} E^v(e^v)^{\top} E(u_n), \\
    e_1^u &= \argmax_{e^u \in {N^u}} E^u(e^u)^{\top} E(u_n),
\end{aligned}
\end{equation}
where $E^v(e^v)$, $E^u(e^u)$ and $E(u_n)$ denote the embeddings for $e^v$, $e^u$ and the last utterance $u_n$ respectively.
The state of the environment is ubiquitous and shared by all agents, but in a multi-agent setting, the observation and the action are both private and only accessible by the individual agent.
Take the video agent as an example. When receiving a local observation $e_t^v$, or the current location on $\mathcal{G}^v$, it will select an outgoing edge $r^v_t$ with its own private policy network $p_v(r|s_{t})$, and obtain a reward from the environment. 
We also design a central communicator to encode historical information and promote multiple agents to work in a collaborative way. The details about the central communicator, the private policy network, and the reward will be described as follows:

\paragraph{Central Communicator.}
To make full use of the information from different modalities and facilitate the reasoning process, we design a central communicator which can get access to the local observations and actions of all agents~\cite{feng2018learning}.
The central communicator works by recursively encoding the historical information~(i.e., $(o_1, a_1, \cdots, o_{t}, a_{t})$) into a message $h_t$ and transporting this message between agents.
Specifically, we implement the central communicator as a recurrent neural network, with the hidden state $h_t$ encoding the past observations and actions.
At time $t$, the central communicator takes current observation $o_t$ and action $a_t$ as input, and updates the hidden state as:
\begin{equation}
\begin{aligned}
h_t         &= \operatorname{RNN}(h_{t-1}, f(a_t, o_t); \phi), \\
f(a_t, o_t) &=W_c[E^v(r^v_t);E^v(e^v_t);E^u(r^u_t);E^u(e^u_t)],
\end{aligned}
\end{equation}
where $W_c$ is a learnable projection matrix, $E^v(r^v_t)$,$E^v(e^v_t)$, $E^u(r^u_t)$ and $E^u(e^u_t)$ are embeddings of $r^v_t$, $e^v_t$, $r^u_t$ and $e^u_t$ respectively.
Consequently, with the help of message $h_{t-1}$, the full state can be approximated as $s^{v}_t \approx (h_{t-1}, e^{v}_{t})$ for the video agent or $s^{v}_u \approx (h_{t-1}, e^{u}_{t})$ for the context agent.

\paragraph{Private Policy Network.}
Each agent has its own private policy network that chooses an outgoing edge at the current location.
Take the video agent as an example.
With the guidance of transported message $h_{t-1}$, the policy network can be approximated as $p_v(r|s_t)\approx p_v(r|h_{t-1}, e^v_t)$, which is formally defined as:
\begin{equation}
\begin{aligned}
    p_v(r|h_{t-1}, e^v_t; \psi^v)&=\frac{e^{g(h_{t-1}, e^v_t, r)}}{\sum_{r \in \mathcal{R}(e^v_t)}e^{g(h_{t-1}, e^v_t, r)}}, \\
    g(h_{t-1},e^v_t, r) &= W_{a}[h_{t-1};E^v(e^v_{t});E^v(r)],
\end{aligned}
\end{equation}
where $W_{a}$ is a learnable parameter and $\mathcal{R}(e^v_t)$ denotes all outgoing edges of the node $e^v_t$.
We define the private policy network for context agent $p_u(r|h_{t-1}, e^u_t;\psi^u)$ following the same procedure.

\paragraph{Reward.}
We only have a reward once the complete chain of reasoning is obtained.
Given the final state $s_T=(o_1, a_1, \cdots, a_{T-1}, o_{T})$
($T$ is the maximum time constraint), we can obtain the reasoning paths $p^v$ and $p^u$ for video and dialogue context respectively, and define the reward as:
\begin{equation}
\begin{aligned}
\label{eq:reward}
    Re(s_T) &= \operatorname{ROUGE}(p^v, p_{gt}) + \operatorname{ROUGE}(p^u, p_{gt}),\\
    p^v&=(e^v_1, r^v_1, \cdots, r^v_{T-1}, e^v_{T}), \\
    p^u&=(e^u_1, r^u_1, \cdots, r^u_{T-1}, e^u_{T}),
\end{aligned}
\end{equation}
where $p_{gt}$ is the subject-predicate-object triplet extracted from the ground-truth response, and $\operatorname{ROUGE}(\cdot, \cdot)$ is a function that returns the ROUGE-1 score~\cite{lin2004rouge} between two sequences.

\subsection{Generation Module}
\label{sec:generation}

We employ the pre-trained GPT-2~\cite{radford2019language} as the backbone of our generation module, which synthesizes a response conditioning on the reasoning path $p^v$ for video data, the reasoning path $p^u$ for dialogue context and the last utterance in context $u_n=\{w_n^1,\cdots,w_n^m\}$.
Formally, the input of the generation module is defined as:
\begin{equation}
    \{e^v_1r^v_1\ldots e^v_T[\mathrm{SEP}]e^u_1r^u_1\ldots e^u_T[\mathrm{SEP}]w^1_n\ldots w^m_n\},
\end{equation}
where $[\mathrm{SEP}]$ is a special token separating different types of data. The probability of generating the response $R=\{w^1_r, w^2_r, \cdots, w^m_r\}$ is formulated as:
\begin{equation}
\begin{aligned}
p(R|V,U;\theta)&\approx p(R|p^v, p^u, u_n; \theta) \\
&=\prod_{i=1}^{m}p(w^{i}_r|w^{<i}_r, p^v, p^u, u_n).
\end{aligned}
\end{equation}

\subsection{Learning Details}
\label{sec:learn}
To estimate $\theta$~(i.e., parameters of the generation module), we directly minimize the negative log-likelihood of response $R$ through MLE loss:
\begin{equation}
    \mathcal{L}_{mle}=-\sum_{i=1}^{m}\log p(w^{i}_r|w^{<i}_r, p^v, p^u, u_n).
\end{equation}

The parameters of private policy networks~(i.e., $\psi^v$ and $\psi^u$), as well as the parameters of the central communicator~(i.e., $\phi$), are optimized through policy-gradient method~\cite{sutton2000policy}. 
Specifically, we sample reasoning paths $\tilde{p}^v$ and $\tilde{p}^u$ according to the private policy networks and the central communicator, and define the loss as follows:
\begin{equation}
\begin{aligned}
\mathcal{L}^v_{pg} &= -Re(\tilde{s}_T)\sum_{t=1}^{T-1}\log p_v(r^v_t|h_{t-1}, e^v_t; \psi^v), \\
\mathcal{L}^u_{pg} &= -Re(\tilde{s}_T)\sum_{t=1}^{T-1}\log p_u(r^u_t|h_{t-1}, e^u_t; \psi^u),
\end{aligned}
\end{equation}
where $Re(\cdot)$ is the reward function defined in Eq.~\ref{eq:reward} and $\tilde{s}_T$ is the final state when sampling the chains of reasoning.
The general training process is conducted by alternately optimizing $\mathcal{L}_{mle}$ and $\mathcal{L}_{pg}=\mathcal{L}^v_{pg}+\mathcal{L}^u_{pg}$.

\begin{table*}[t!]
\centering
\small
\resizebox{0.8\linewidth}{!}{
\begin{tabular}{lccccccc}
\toprule
\multicolumn{1}{l}{Model}           & \multicolumn{1}{c}{BLEU1} & \multicolumn{1}{c}{BLEU2} & \multicolumn{1}{c}{BLEU3} & \multicolumn{1}{c}{BLEU4} & \multicolumn{1}{c}{METEOR} & \multicolumn{1}{c}{ROUGE-L} & CIDEr \\ \midrule
\multicolumn{8}{c}{\textit{with} caption}                                                                                                                                                                                    \\ \midrule
\multicolumn{1}{l}{Naive Fusion}    & \multicolumn{1}{c}{0.628}     & \multicolumn{1}{c}{0.481}     & \multicolumn{1}{c}{0.377}     & \multicolumn{1}{c}{0.298}     & \multicolumn{1}{c}{0.220}      & \multicolumn{1}{c}{0.491}       & 0.748     \\ 
\multicolumn{1}{l}{MTN}             & \multicolumn{1}{c}{0.731}     & \multicolumn{1}{c}{0.597}     & \multicolumn{1}{c}{0.494}     & \multicolumn{1}{c}{0.410}     & \multicolumn{1}{c}{0.274}      & \multicolumn{1}{c}{0.569}       & 1.129     \\ 
\multicolumn{1}{l}{Student-Teacher} & \multicolumn{1}{c}{0.727}     & \multicolumn{1}{c}{0.593}     & \multicolumn{1}{c}{0.488}     & \multicolumn{1}{c}{0.405}     & \multicolumn{1}{c}{0.273}      & \multicolumn{1}{c}{0.566}       & 1.118     \\ 
\multicolumn{1}{l}{RLM}             & \multicolumn{1}{c}{0.765}     & \multicolumn{1}{c}{0.643}     & \multicolumn{1}{c}{0.543}     & \multicolumn{1}{c}{0.459}     & \multicolumn{1}{c}{0.294}      & \multicolumn{1}{c}{0.606}       & 1.308     \\ 
\multicolumn{1}{l}{VGD-GPT2}        & \multicolumn{1}{c}{0.750}     & \multicolumn{1}{c}{0.621}     & \multicolumn{1}{c}{0.516}     & \multicolumn{1}{c}{0.433}     & \multicolumn{1}{c}{0.283}      & \multicolumn{1}{c}{0.581}       & 1.196     \\ 
\multicolumn{1}{l}{BiST}            & \multicolumn{1}{c}{0.755}     & \multicolumn{1}{c}{0.619}     & \multicolumn{1}{c}{0.510}     & \multicolumn{1}{c}{0.429}     & \multicolumn{1}{c}{0.284}      & \multicolumn{1}{c}{0.581}       & 1.192     \\ 
\multicolumn{1}{l}{PDC-GPT}         & \multicolumn{1}{c}{0.770}     & \multicolumn{1}{c}{\textbf{0.653}}     & \multicolumn{1}{c}{0.539}     & \multicolumn{1}{c}{0.449}     & \multicolumn{1}{c}{0.292}      & \multicolumn{1}{c}{0.606}       & 1.295     \\ 
\multicolumn{1}{l}{Ours}            & \multicolumn{1}{c}{\textbf{0.776}$^{\star}$}     & \multicolumn{1}{c}{0.652}     & \multicolumn{1}{c}{\textbf{0.551}$^{\star}$}     & \multicolumn{1}{c}{\textbf{0.466}$^{\star}$}     & \multicolumn{1}{c}{\textbf{0.304}$^{\star}$}      & \multicolumn{1}{c}{\textbf{0.609}}       & \textbf{1.333}$^{\star}$     \\ \midrule
\multicolumn{8}{c}{\textit{without} caption}                                                                                                                                                                                 \\ \midrule
\multicolumn{1}{l}{Naive Fusion}    & \multicolumn{1}{c}{0.626}     & \multicolumn{1}{c}{0.485}     & \multicolumn{1}{c}{0.383}     & \multicolumn{1}{c}{0.309}     & \multicolumn{1}{c}{0.251}      & \multicolumn{1}{c}{0.487}       & 0.746     \\ 
\multicolumn{1}{l}{MTN}             & \multicolumn{1}{c}{0.692}     & \multicolumn{1}{c}{0.556}     & \multicolumn{1}{c}{0.459}     & \multicolumn{1}{c}{0.368}     & \multicolumn{1}{c}{0.259}      & \multicolumn{1}{c}{0.537}       & 0.964     \\ 
\multicolumn{1}{l}{Student-Teacher} & \multicolumn{1}{c}{0.675}     & \multicolumn{1}{c}{0.543}     & \multicolumn{1}{c}{0.446}     & \multicolumn{1}{c}{0.371}     & \multicolumn{1}{c}{0.248}      & \multicolumn{1}{c}{0.527}       & 0.966     \\ 
\multicolumn{1}{l}{RLM}             & \multicolumn{1}{c}{0.694}     & \multicolumn{1}{c}{0.570}     & \multicolumn{1}{c}{0.476}     & \multicolumn{1}{c}{0.402}     & \multicolumn{1}{c}{0.254}      & \multicolumn{1}{c}{0.544}       & 1.052     \\ 
\multicolumn{1}{l}{VGD-GPT2}        & \multicolumn{1}{c}{0.692}     & \multicolumn{1}{c}{0.563}     & \multicolumn{1}{c}{0.464}     & \multicolumn{1}{c}{0.388}     & \multicolumn{1}{c}{0.251}      & \multicolumn{1}{c}{0.539}       & 0.998     \\ 
\multicolumn{1}{l}{BiST}            & \multicolumn{1}{c}{0.715}     & \multicolumn{1}{c}{0.560}     & \multicolumn{1}{c}{0.477}     & \multicolumn{1}{c}{0.390}     & \multicolumn{1}{c}{0.259}      & \multicolumn{1}{c}{0.552}       & 1.030     \\ 
\multicolumn{1}{l}{PDC-GPT}         & \multicolumn{1}{c}{0.713}     & \multicolumn{1}{c}{0.574}     & \multicolumn{1}{c}{0.468}     & \multicolumn{1}{c}{0.385}     & \multicolumn{1}{c}{0.260}      & \multicolumn{1}{c}{0.545}       & 1.010     \\
\multicolumn{1}{l}{Ours}            & \multicolumn{1}{c}{\textbf{0.717}}     & \multicolumn{1}{c}{\textbf{0.590}$^{\star}$}     & \multicolumn{1}{c}{\textbf{0.491}$^{\star}$}     & \multicolumn{1}{c}{\textbf{0.414}$^{\star}$}     & \multicolumn{1}{c}{\textbf{0.265}$^{\star}$}      & \multicolumn{1}{c}{\textbf{0.558}$^{\star}$}       & \textbf{1.078}$^{\star}$     \\ \bottomrule
\end{tabular}
}
\caption{Automatic evaluation results on the test set of AVSD$@$DSTC7. Numbers in bold are the best results. Significant improvements over the best baseline results are marked with $\star$ (t-test with p-value $<0.05$).}
\vspace{-2mm}
\label{tab:exp-dstc}
\end{table*}

\section{Experiments}
\subsection{Datasets}
We evaluate our model on two benchmark datasets for video-grounded dialogue generation: 
\paragraph{AVSD$@$DSTC7}
This dataset is constructed by ~\citet{alamri2018audio} through crowd-sourcing and contains conversations about Charades videos~\cite{sigurdsson2016hollywood}. 
\paragraph{Twitch-FIFA.}
This dataset is collected by crawling live-broadcast soccer game videos and the chats from Twitch.tv~\cite{pasunuru2018game}. 

To facilitate reproducibility, we adopt the datasets shared by the publishers and conduct pre-processing strictly following the official code. Table~\ref{tab:stat} reports the statistics of AVSD$@$DSTC7 and Twitch-FIFA.

\begin{table}[!ht]
\centering
\resizebox{1\linewidth}{!}{
\begin{tabular}{lcccccc}
\toprule
\multirow{2}{*}{} & \multicolumn{3}{c}{AVSD$@$DSTC7} & \multicolumn{3}{c}{Twitch-FIFA} \\ \cmidrule(lr){2-4}\cmidrule(lr){5-7}
                            & Train   & Valid     & Test  & Train     & Valid     & Test     \\ \midrule
Number of Dialogues                  & 7,659   & 1,787   & 1,710 & 10,150         & 2,153       & 2,780        \\ 
Number of Utterances                    & 76,590  & 17,870  & 6,745 & 110,602         & 19,362       & 31,245        \\ 
Average Turns per Dialogue         & 10      & 10      & 3.94  & 10.52         & 8.99       & 11.24        \\ 
\bottomrule
\end{tabular}
}
\caption{Statistics of the two datasets.}
\vspace{-2mm}
\label{tab:stat}
\end{table}

\subsection{Evaluation Metrics}
\paragraph{Automatic Evaluation.}
We choose $4$ commonly used reference-based metrics including BLEU~\cite{papineni2002bleu}, ROUGE~\cite{lin2004rouge}, METEOR~\cite{lavie2007meteor} and CIDEr~\cite{vedantam2015cider}.
We evaluate our models
using the official code released by the owner of AVSD$@$DSTC7 dataset~\footnote{\scriptsize\url{https://drive.google.com/open?id=1nz9Pu9YIfuZHzowhASXERajRXqE6DBQx}}.

\begin{table*}[t!]
\centering
\small
\resizebox{0.75\linewidth}{!}{
\begin{tabular}{lccccccc}
\toprule
\multicolumn{1}{l}{Model}           & \multicolumn{1}{c}{BLEU1} & \multicolumn{1}{c}{BLEU2} & \multicolumn{1}{c}{BLEU3} & \multicolumn{1}{c}{BLEU4} & \multicolumn{1}{c}{METEOR} & \multicolumn{1}{c}{ROUGE-L} & CIDEr \\ \midrule
\multicolumn{1}{l}{BIDAF}    & \multicolumn{1}{c}{0.092}     & \multicolumn{1}{c}{0.057}     & \multicolumn{1}{c}{0.043}     & \multicolumn{1}{c}{0.035}     & \multicolumn{1}{c}{0.036}      & \multicolumn{1}{c}{0.134}       & 0.099    \\ 
\multicolumn{1}{l}{MTN}             & \multicolumn{1}{c}{0.113}     & \multicolumn{1}{c}{0.06}     & \multicolumn{1}{c}{0.043}     & \multicolumn{1}{c}{0.034}     & \multicolumn{1}{c}{0.039}      & \multicolumn{1}{c}{0.143}       & 0.091     \\ 
\multicolumn{1}{l}{BiST} & \multicolumn{1}{c}{0.100}     & \multicolumn{1}{c}{0.050}     & \multicolumn{1}{c}{0.031}     & \multicolumn{1}{c}{0.022}     & \multicolumn{1}{c}{0.038}      & \multicolumn{1}{c}{0.159}       & 0.104     \\ 
\multicolumn{1}{l}{RLM}             & \multicolumn{1}{c}{0.102}     & \multicolumn{1}{c}{0.081}     & \multicolumn{1}{c}{0.070}     & \multicolumn{1}{c}{0.060}     & \multicolumn{1}{c}{0.046}      & \multicolumn{1}{c}{0.188}       & 0.130     \\ 
\multicolumn{1}{l}{Ours}        & \multicolumn{1}{c}{\textbf{0.128}$^{\star}$}     & \multicolumn{1}{c}{\textbf{0.101}$^{\star}$}     & \multicolumn{1}{c}{\textbf{0.083}$^{\star}$}     & \multicolumn{1}{c}{\textbf{0.069}$^{\star}$}     & \multicolumn{1}{c}{\textbf{0.052}$^{\star}$}      & \multicolumn{1}{c}{\textbf{0.193}}       & \textbf{0.176}$^{\star}$     \\ \bottomrule
\end{tabular}
}
\caption{Automatic evaluation results on the test set of Twitch-FIFA. Numbers in bold are the best results. Significant improvements over the best baseline results are marked with $\star$ (t-test with p-value $<0.05$).}
\vspace{-2mm}
\label{tab:exp-fifa}
\end{table*}

\paragraph{Human Evaluation.}
We also conduct a human evaluation to deepen our understanding of the quality of responses produced by different models.
We randomly sample $300$ examples from the test sets of AVSD$@$DSTC7, and hire $6$ well-educated native speakers to conduct qualitative analysis on the results produced by our model and all competitive baselines, which are randomly mixed to obscure identification.
The annotators evaluate the quality of the responses using three criteria: 
(1) \emph{Language Fluency}: whether the response is fluent and devoid of grammatical errors,
(2) \emph{Context Coherence}: whether the response is coherent with the dialogue context,
and (3) \emph{Factual Correctness}: whether the response is factually consistent with the events depicted in the video.
Each annotator rates each response for each aspect with a score from $\{0, 1, 2\}$ (representing ``bad'', ``fair'' and ``good'' respectively).
Each response receives three scores for the aforementioned $3$ aspects, and Fleiss' kappa~\cite{fleiss1971measuring} is used to gauge the level of agreement between all annotators.

\subsection{Baseline Models}
The following models are selected as baselines:
(1) \textbf{Naive Fusion:} A model proposed by \citet{hori2019end} which combines all modalities with a projection matrix.
(2) \textbf{MTN:} A model proposed by \citet{Le2019MultimodalTN} that is based on transformer architecture and employs query-guided attention to extract query-aware features from videos.
(3) \textbf{Student-Teacher:} A model proposed by~\citet{hori2019joint} that aims to alleviate the reliance on human-generated captions through a student-teaching learning method.
(4) \textbf{RLM:} A model proposed by ~\citet{li2020bridging} that is trained with multi-task learning objectives to learn joint representations among different modalities.
(5) \textbf{VGD-GPT2:} A model proposed by \citet{le2020video} that leverages the power of pre-trained language models for improving video-grounded dialogue.
(6) \textbf{BiST:} A model proposed by \citet{le2020bist} that exploits both spatial and temporal-level information to promote video understanding.
(7) \textbf{PDC+GPT2:} A model proposed by \citet{le2021learning} that conducts reasoning on the dialogue history to model the information flow at turn level.

All the baselines are taken from their open-source implementations or re-implemented strictly following the details in the original papers.

\subsection{Implementation Details}
In our experiments, the maximum time constraint $T$ which serves as the stop criteria for reasoning is set as $3$.
The embedding sizes for relations and entities are all set as $100$.
The central communicator is implemented as an LSTM network whose size of the hidden state is set as $200$.
The generation module is implemented on the basis
of the pre-trained GPT-2~(small) model which has $117M$ parameters.
All models are learned with Adam~\cite{kingma2014adam} optimizer with $\beta_1=0.9$ and $\beta_2=0.999$.
We initialize the learning rates to $0.001$ and $6.25e-5$ for the multi-modal reasoning module and the generation module respectively and optimize the model with a linear learning rate decay strategy.
The batch size is set as $8$ in our experiments.
In the test phase, we employ beam search in response decoding and set the beam size, max decode length and the length penalty as $4$, $16$ and $0.1$ respectively.
Early stopping on validation is adopted as a regularization
strategy. All models are trained on an $8\times$RTX
3090 Ti machine.

\begin{table}[t!]
\centering
\resizebox{\linewidth}{!}{
\begin{tabular}{lcccc}
\toprule
\multicolumn{1}{c}{Model} & \begin{tabular}[c]{@{}c@{}}Language\\ Fluency\end{tabular} & \begin{tabular}[c]{@{}c@{}}Context\\ Coherence\end{tabular} & \begin{tabular}[c]{@{}c@{}}Factual\\ Correctness\end{tabular} & Kappa \\ \midrule
RLM                           & 1.73                                                       & 1.67                                                        & 1.47                                                          & 0.61  \\ 
VGD-GPT2                      & 1.76                                                       & 1.54                                                        & 1.50                                                          & 0.65  \\
BiST                          & 1.64                                                       & 1.58                                                        & 1.49                                                          & 0.71  \\ 
PDC-GPT                       & 1.78                                                       & 1.68                                                        & 1.53                                                          & 0.64  \\ \midrule
Ours                          & \textbf{1.81}                                                       & \textbf{1.75}                                                        & \textbf{1.65}                                                          & 0.76  \\ 
\bottomrule
\end{tabular}
}
\caption{Human evaluation results on AVSD$@$DSTC7. Numbers in bold are the best results.}
\vspace{-2mm}
\label{tab:exp-human}
\end{table}

\begin{table*}[t!]
\centering
\small
\resizebox{0.8\linewidth}{!}{
\begin{tabular}{lccccccc}
\toprule
\multicolumn{1}{c}{Model}           & \multicolumn{1}{c}{BLEU1} & \multicolumn{1}{c}{BLEU2} & \multicolumn{1}{c}{BLEU3} & \multicolumn{1}{c}{BLEU4} & \multicolumn{1}{c}{METEOR} & \multicolumn{1}{c}{ROUGE-L} & CIDEr \\ \midrule
\multicolumn{1}{c}{Ours}        & \multicolumn{1}{c}{0.717}     & \multicolumn{1}{c}{0.590}     & \multicolumn{1}{c}{0.491}     & \multicolumn{1}{c}{0.414}     & \multicolumn{1}{c}{0.265}      & \multicolumn{1}{c}{0.558}       & 1.078    \\ 
\midrule
\multicolumn{1}{c}{\textit{-$\mathcal{G}^{u}$}} & \multicolumn{1}{c}{0.706}     & \multicolumn{1}{c}{0.578}     & \multicolumn{1}{c}{0.481}     & \multicolumn{1}{c}{0.403}     & \multicolumn{1}{c}{0.261}      & \multicolumn{1}{c}{0.554}       & 1.067     \\ 
\multicolumn{1}{c}{\textit{-$\mathcal{G}^{v}$}}             & \multicolumn{1}{c}{0.704}     & \multicolumn{1}{c}{0.579}     & \multicolumn{1}{c}{0.481}     & \multicolumn{1}{c}{0.405}     & \multicolumn{1}{c}{0.260}      & \multicolumn{1}{c}{0.552}       & 1.063     \\ 
\multicolumn{1}{c}{\textit{-$\mathcal{G}^u$ \& $\mathcal{G}^v$}}    & \multicolumn{1}{c}{0.697}     & \multicolumn{1}{c}{0.573}     & \multicolumn{1}{c}{0.474}     & \multicolumn{1}{c}{0.399}     & \multicolumn{1}{c}{0.257}      & \multicolumn{1}{c}{0.545}       & 1.048    \\ 
\multicolumn{1}{c}{\textit{-Communicator}}        & \multicolumn{1}{c}{0.708}     & \multicolumn{1}{c}{0.578}     & \multicolumn{1}{c}{0.478}     & \multicolumn{1}{c}{0.400}     & \multicolumn{1}{c}{0.259}      & \multicolumn{1}{c}{0.548}       & 1.058     \\ 
\bottomrule
\end{tabular}
}
\caption{Ablation study on AVSD$@$DSTC7. All experiments are conducted in the \emph{without caption} setting.}
\label{tab:exp-abl}
\end{table*}

\begin{table}[t!]
\centering
\setlength{\tabcolsep}{1.5pt}
\resizebox{\linewidth}{!}{
\begin{tabular}{cccccccc}
\toprule
Model & BLEU1-4 & METEOR & ROUGE-L & CIDEr \\ \midrule
T=1 & 0.713/0.584/0.485/0.407 & 0.262 & 0.551 & 1.055 \\
T=2 & 0.717/0.590/0.491/0.414 & 0.265 & 0.558 & 1.078 \\
T=3 & 0.714/0.588/0.489/0.412 & 0.265 & 0.554 & 1.082 \\
\bottomrule
\end{tabular}
}
\caption{Performance of our model under different maximum time constraints.}
\vspace{-2mm}
\label{tab:exp-time}
\end{table}

\begin{figure*}[t!]
\centering
\includegraphics[width=0.88\textwidth]{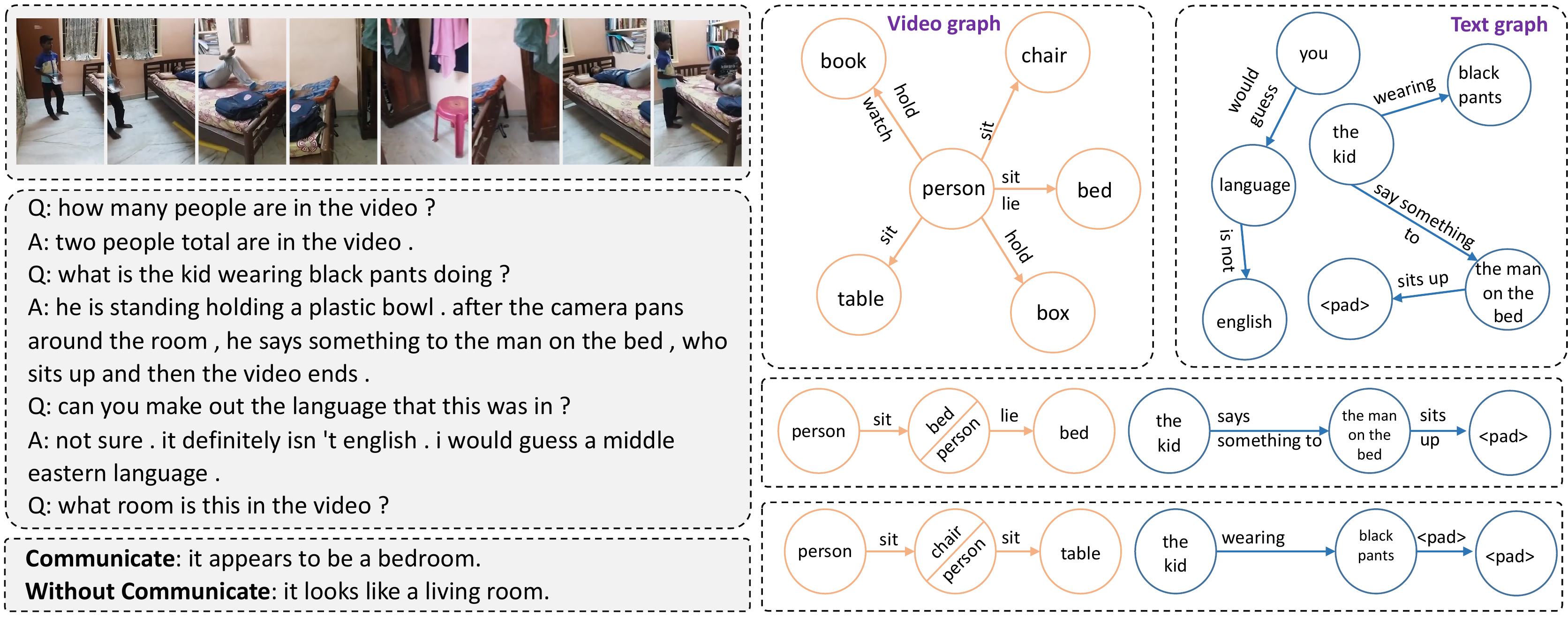}
\caption{A case from the test set of AVSD$@$DSTC7.}
\label{fig:case}
\end{figure*}

\subsection{Evaluation Results}
In this section, we will compare the performance of various models on AVSD$@$DSTC7 and Twitch-FIFA. 
We conduct two experiment settings for AVSD$@$DSTC7, including \emph{with caption} and \emph{without caption}, since the video caption is unavailable in most real-world scenarios.
Table~\ref{tab:exp-dstc} and Table~\ref{tab:exp-fifa} show the performance of our model on AVSD$@$DSTC and Twitch-FIFA respectively. From the results, we can observe that:

(1) Our model achieves the new state-of-the-art on most metrics in both datasets, illuminating the effectiveness of the proposed multi-agent reasoning framework and the multi-modal semantic graphs.
In particular, the proposed model outperforms RLM and PDC-GPT, the two best baselines on AVSD$@$DSTC7, since they both directly feed the video features for the generation procedure which presents obstacles for the PLMs to conduct multi-modal reasoning.
This is also supported by the results in our pilot study~(as shown in Table~\ref{tab:exp-pilot}).

(2) In the AVSD$@$DSTC7 dataset, the caption has a significant impact on models since, in the absence of caption, all models' performance significantly degrades. Another intriguing finding is that reasoning-based methods~(e.g., BiST, PDC-GPT and ours) rely less on the video caption as compared to methods without explicit reasoning~(e.g., RLM and VGD-GPT2). 
This confirms the need for multi-modal reasoning, and our proposed method of collaborative reasoning is more effective.

(3) Because there is a lot of noise in the live-broadcast data collection, which is closer to the real-world scenario than the manually-labeled dataset, the PLMs-based methods~(e.g., RLM and ours) perform better on Twitch-FIFA than others. This further emphasizes the value of integrating PLMs with multi-modal reasoning, one of the benefits of our proposed method.

\paragraph{Human Evaluation.}
Table~\ref{tab:exp-human} shows the results of human evaluation. 
Although our model achieves a language fluency score that is comparable to other baselines, it attains a significant improvement in context coherency and factual correctness, which is congruent with the results of our pilot experiments and automatic evaluation.
The fact that all kappa values are more than $0.6$ shows that the annotators are in agreement.

\subsection{Discussions}

\paragraph{Ablation Study.}
In addition to the main experiments, we compare the full model with the following variations to gain a better understanding of how each component affects the general performance:
(1) \textit{-$\mathcal{G}^{u}$}: the context graph is removed;
(2) \textit{-$\mathcal{G}^{v}$}: the video graph is removed;
(3) \textit{-$\mathcal{G}^u$ \& $\mathcal{G}^v$}: both the context graph and the video graph are removed. Here, the model directly generates the response based on the dialogue context and the video features provided by \citet{alamri2018audio};
and (4) \textit{-Communicator}: the Central Communicator is removed. In this instance, the context agent and video agent each independently reason on graphs.
The experiment results of ablation are shown in Table~\ref{tab:exp-abl}.
We can draw the following conclusions: (1) the multi-modal semantic graphs are both significant, as the performance is negatively impacted by deleting one or more of them. Although they are built using off-the-shelf tools with heuristics, they nonetheless contain significant information that enables the agents to locate chains of reasoning that lead to solutions;
and (2) the communicator is helpful because it enables crucial signals from different modalities to reinforce each other.

\paragraph{Effect the Maximum Time Constraint $T$.}
We continue to look at how sensitive the model is to various selections of the maximum time constraint $T$. 
In order to achieve this, we change the value of $T$ in $\{1; 2; 3\}$, and then report the evaluation results in Table~\ref{tab:exp-time}.
As can be shown, our model performs best when $T=2$ since a larger maximum time constraint will introduce more irrelevant entities and relations into generation, whereas a smaller number (i.e., $T=1$) limits the reasoning paths to only the entity that is most similar to the last utterance.

\paragraph{Case Study.}
We further conduct a case study to have a deeper understanding of the multi-modal reasoning process in our model.
Figure~\ref{fig:case} shows an example from the test set of AVSD@DSTC7.
We can see that our model is able to precisely construct the reasoning paths for dialogue context and video respectively, and to produce a response that accurately captures the factual information in the video.
For comparison, we also provide the results of a variant in which the central communicator has been eliminated.
We can observe that the communication mechanism can effectively assist in retrieving relevant signals from multi-modal data.
\section{Conclusion}

We propose a multi-modal reasoning framework that can be used in conjunction with PLMs to enable the complementation of information from various modalities.
Specifically, we devise a video agent and a context agent to extract reasoning paths on video and dialogue contexts respectively.
A central communicator is also designed to transport information between the two agents and enables their cooperative operation.
The general framework is optimized through multi-agent reinforcement learning.
Evaluation results on two benchmarks indicate that our model can significantly outperform state-of-the-art methods.

\section*{Limitations}
We also recognize that our model has its certain limitations: (i) Due to multi-modal semantic graphs, our framework needs higher computation overheads to extract triplet relations from video and perform reasoning on dual graphs. Nonetheless, the multi-modal reasoning paths which are compatible with PLMs make our model still practical and scalable. (ii) The performance of our model may be limited to some extent by the quality of the dual graphs created by off-the-shelf tools. 

\section*{Ethics Statement}
This paper studies video-grounded dialogue generation and proposes a multi-modal reasoning framework based on multi-agent reinforcement learning to facilitate the complementarity of information from different modalities. There are no ethical concerns with this study. The datasets we used are widely used by other academics and are typically accessible to the public. No ethical or societal prejudice is introduced by the suggested strategy.

\section*{Acknowledgements}
We appreciate the anonymous reviewers for their constructive comments. This work was supported by the National Key Research and Development Program of China (No. 2020AAA0106600).

\bibliography{anthology,custom}
\bibliographystyle{acl_natbib}

\end{document}